\documentclass[10pt,twocolumn,letterpaper]{article}

\usepackage{cvpr}
\usepackage{times}
\usepackage{epsfig}
\usepackage{graphicx}
\usepackage{amsmath}
\usepackage{amssymb}
\usepackage{subfigure}
\usepackage{authblk}

\usepackage[breaklinks=true,bookmarks=false,colorlinks]{hyperref}
\cvprfinalcopy % *** Uncomment this line for the final submission

\def\cvprPaperID{592} % *** Enter the CVPR Paper ID here

\ifcvprfinal\fi
\begin{document}

%%%%%%%%% TITLE
\title{Gradually Vanishing Bridge for Adversarial Domain Adaptation}

\author{\vspace{-0.2in}Shuhao Cui$^{1,2}$ ~~ Shuhui Wang$^{1}$\thanks{Corresponding author.} ~~ Junbao Zhuo$^{1,2}$ ~~ Chi Su$^{3}$ ~~ Qingming Huang$^{1,2,4}$ ~~ Qi Tian$^{5}$\\
	$^{1}$Key Lab of Intell. Info. Process., Inst. of Comput. Tech., CAS, Beijing, China\\
	$^{2}$University of Chinese Academy of Sciences, Beijing, China ~~ $^{3}$Kingsoft Cloud, Beijing, China\\
	$^{4}$Peng Cheng Laboratory, Shenzhen, China ~~  $^{5}$Noah's Ark Lab, Huawei Technologies\\
	{\tt\small \{cuishuhao18s, wangshuhui\}@ict.ac.cn, junbao.zhuo@vipl.ict.ac.cn, suchi@kingsoft.com, qmhuang@ucas.ac.cn, tian.qi1@huawei.com\vspace{-0.15in}}}

\maketitle
\thispagestyle{empty}
%\vspace{-10pt}
%%%%%%%%% ABSTRACT
\begin{abstract}
\vspace{-5pt}
In unsupervised domain adaptation, rich domain-specific characteristics bring great challenge to learn domain-invariant representations. However, domain discrepancy is considered to be directly minimized in existing solutions, which is difficult to achieve in practice. Some methods alleviate the difficulty by explicitly modeling domain-invariant and domain-specific parts in the representations, but the adverse influence of the explicit construction lies in the residual domain-specific characteristics in the constructed domain-invariant representations. In this paper, we equip adversarial domain adaptation with Gradually Vanishing Bridge (GVB) mechanism on both generator and discriminator. On the generator, GVB could not only reduce the overall transfer difficulty, but also reduce the influence of the residual domain-specific characteristics in domain-invariant representations. On the discriminator, GVB contributes to enhance the discriminating ability, and balance the adversarial training process. Experiments on three challenging datasets show that our GVB methods outperform strong competitors, and cooperate well with other adversarial methods. The code is available at \url{https://github.com/cuishuhao/GVB}.
\end{abstract}

\vspace{-5pt}
%%%%%%%%% BODY TEXT
\section{Introduction}

Diverse vision applications have gained significant improvement with the cutting-edge technologies such as deep convolutional neural networks (CNN). Despite of success already achieved, even a subtle departure from training data may still cause existing shallow or deep models to make spurious predictions. Directly applying a deep CNN well-trained on millions of images to a new domain encounters performance degradation, while collecting labeled samples for the new domain is expensive and time-consuming. To alleviate such notorious domain discrepancy, researchers resort to unsupervised domain adaptation (UDA)~\cite{chen2017no,gopalan2011domain,sohn2017unsupervised,tzeng2015simultaneous,zhang2017curriculum,zhuo2019unsupervised} that transfers knowledge from labeled source data to related unlabeled target data.

\begin{figure}[t]
	\begin{center}
		%#fbox{\rule{0pt}{2in} \rule{0.9\linewidth}{0pt}}
		\includegraphics[width=0.98\linewidth]{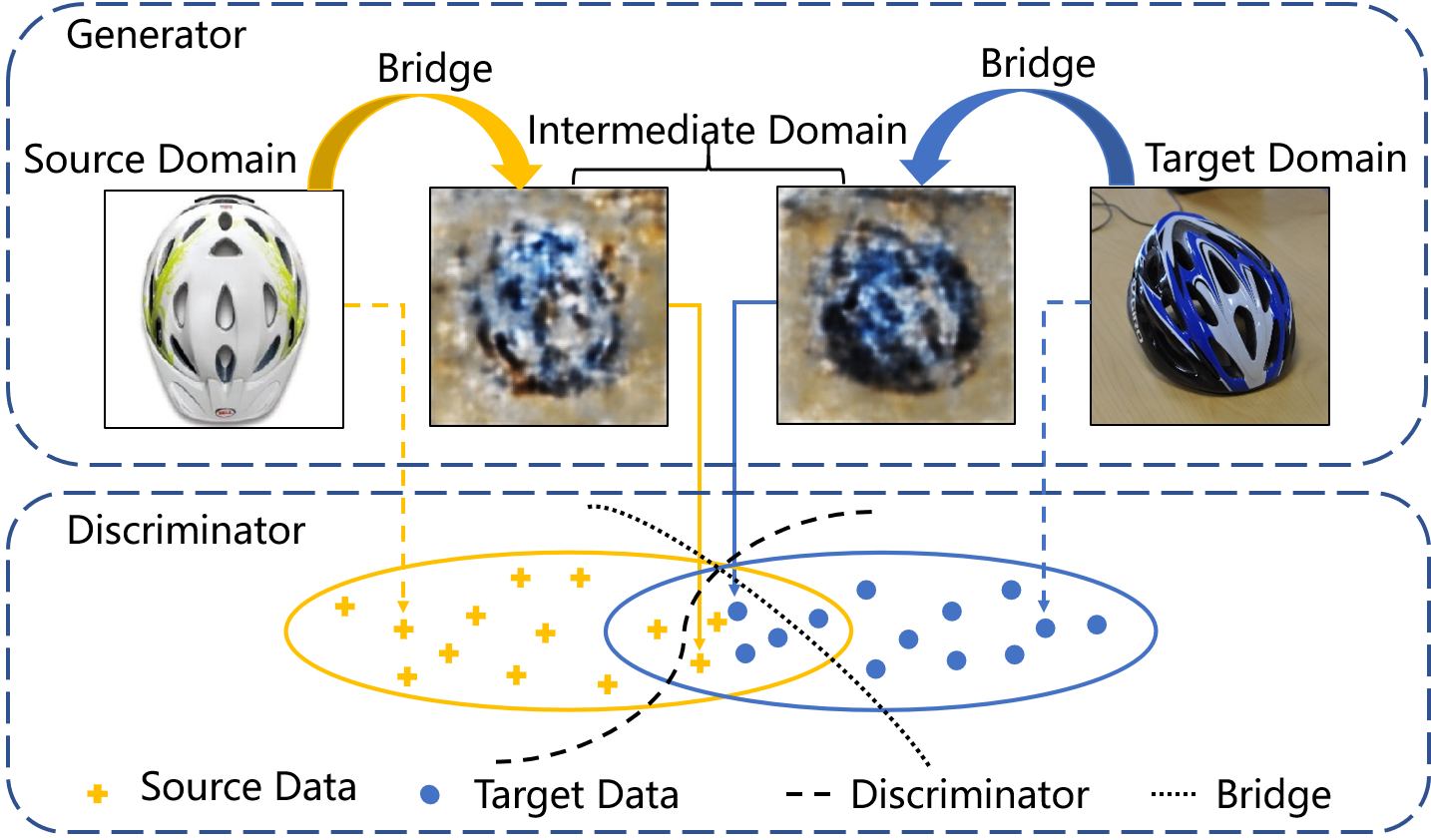}
	\end{center}
	\vspace{-1ex}
	\caption{Illustration of the bridge for adversarial domain adaptation. On the generator, the bridge models domain-specific representations and connects either source or target domain to intermediate domain. On the discriminator, the bridge balances the adversarial game by providing additive adjustable discriminating ability. In these processes, the method to construct bridge is the key issue in our study.}
	\label{softmatching}
	\vspace{-1ex}
\end{figure}

In traditional UDA methods, domain discrepancy is reduced by learning transferable representations~\cite{gong2012geodesic, gretton2012kernel,ni2013subspace}, or reweighing the importance of the samples~\cite{pan2011domain}. Deep methods carry forward the thought of learning transferable representations by distribution alignment on multiple layers~\cite{ganin2014unsupervised,ganin2016domain,long2015learning,yan2017mind,zhuo2017deep}.
Recently, many domain adaptation methods are inspired by Generative Adversarial Networks~(GAN)~\cite{arjovsky2017wasserstein,goodfellow2014generative,mirza2014conditional,zhu2017unpaired}. Based on GAN, the transferable representations are constructed by the generator that tries to fool the discriminator in the {\it minmax} game~\cite{ganin2016domain,hoffman2017cycada,saito2018maximum,sankaranarayanan2018generate,tzeng2017adversarial}. In these studies, the discrepancy between source and target domains is considered to be {\it directly} minimized. Towards more robust UDA, we reinvestigate the intrinsic weakness possessed by existing domain discrepancy minimization paradigms.

To reduce domain discrepancy, the great challenge lies in the rich domain-specific characteristics, which are deemed to be mitigated by distribution alignment~\cite{ganin2014unsupervised,long2015learning} or adversarial training~\cite{ganin2016domain,saito2018maximum}. Towards more comprehensive reduction of domain-specific characteristics, some methods~\cite{bousmalis2016domain,chang2019all,gong2019dlow} explicitly model domain-specific and domain-invariant representations by input image reconstruction. The time-consuming reconstruction function demands rich domain characteristics in domain-specific representations. Such characteristics inevitably result in more domain properties in the intermediate domain, which is expected to be domain-invariant. 
From another perspective of adversarial learning, the adversarial minmax game could achieve better results by balancing the learning ability between the generator and discriminator. Strengthening the discriminating ability by multiple discriminators~\cite{pei2018multi} may facilitate better collaboration with a strong generator in some situations, while the overly critical multiple discriminators may also break the brittal balance of adversarial training.

In this paper, we define the bridge as a basic concept, which is a measurement modeling the discrepancy between the existing and ideal representations. As shown by example in Figure~\ref{softmatching}, the bridge is applied to the generator and discriminator. The bridge on generator models domain-specific parts and connects both source and target domains to the intermediate domain, thus it could mitigate the overall transfer difficulty and facilitate more comprehensive domain alignment. The bridge on discriminator measures the discrepancy between the discriminating function hyperplane and the ideal domain decision boundary.

The key constraint on our constructed bridge is that the range of the bridge is progressively reduced. With the constraint, the bridge mechanism is called Gradually Vanishing Bridge (GVB) for adversarial domain adaptation. We denote GVB applied to generator and discriminator as GVB-G and GVB-D, respectively. Among the bridges, the bridge on generator models domain-specific parts, thus GVB-G could explicitly reduce domain-specific characteristics. Therefore, GVB-G plays the main role in learning domain-invariant representations. It could also reduce the adverse influence of domain characteristics in the representations, and avoid the influence of hard examples with excessive domain characteristics.

Both GVB-G and GVB-D are integrated into the whole framework, denoted as GVB-GD, to ensure the balance and robustness of the two-player minmax game. Experiments show that GVB-GD outperforms competitors on three challenging datasets and achieves state-of-the-art on Office-Home. We also apply GVB to other UDA methods including CDAN~\cite{long2018conditional} and Symnets~\cite{zhang2019domain}. The improvements on CDAN and Symnets show the general applicability of GVB on adversarial domain adaptation. For better explanation, we visualize the bridge in GVB-GD to validate that the bridge could measure the domain-specific characteristics. We further verify the functionality and necessity of GVB, by observation that larger range of GVB-G output tends to result in higher misclassification probability.

In summary, the key contributions are as follows:
\begin{itemize}
	\item We propose a new framework which constructs GVB on both generator and discriminator to achieve more balanced adversarial training for domain adaptation.
	\item As the key role in the framework, GVB on generator could alleviate the transfer difficulty and explicitly reduce the domain characteristics in the representations.
	\item GVB outperforms competitors in most cases, and visualization also show the positive effect of GVB. 
\end{itemize}

\section{Related Work}
\begin{figure*}
	\begin{center}
		%\fbox{\rule{0pt}{2in} \rule{.9\linewidth}{0pt}}
		\includegraphics[width=.96\textwidth]{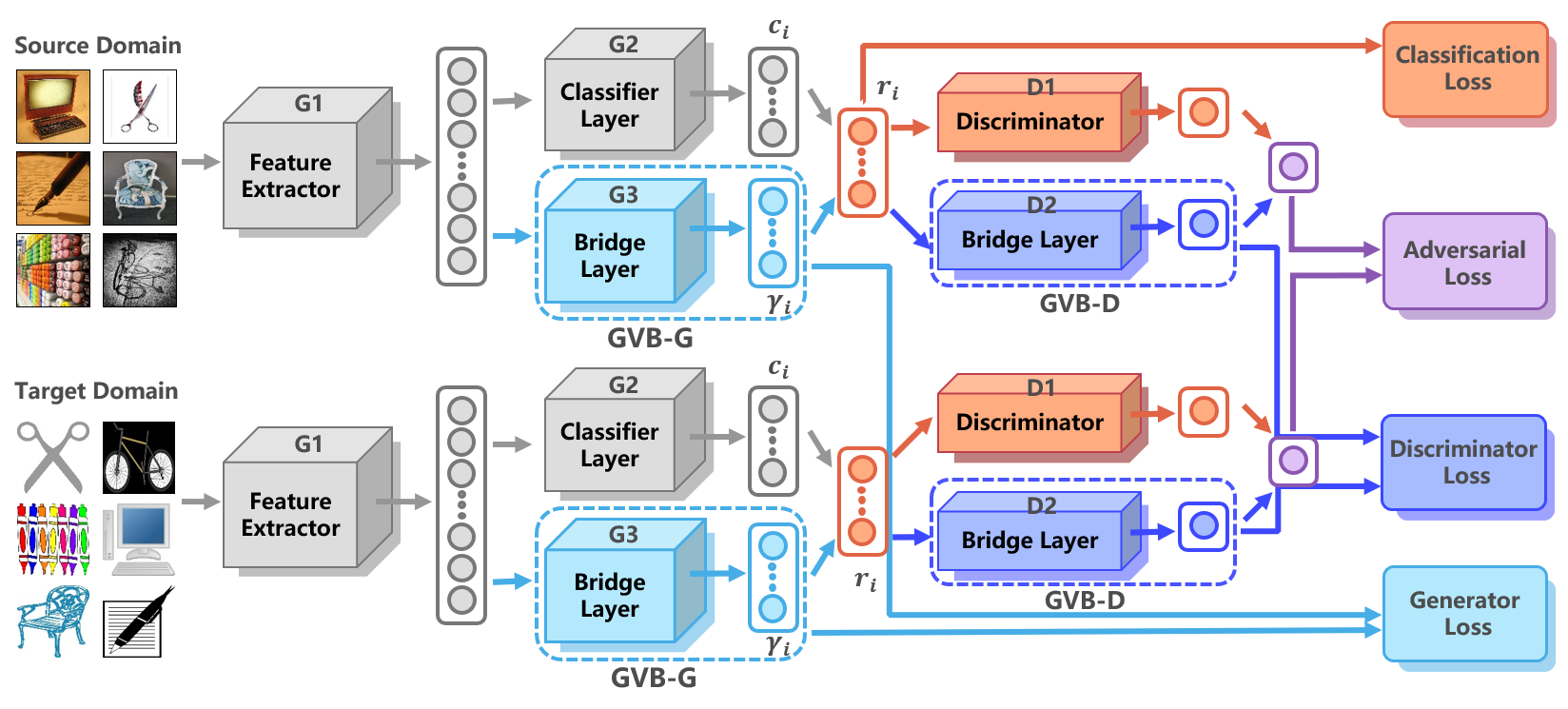}
	\end{center}
	\vspace{-2ex}
	\caption{The proposed GVB-GD framework, which is an end-to-end network shared by source domain and target domain. In GVB-G, the bridge layer $G_3$ outputs $\gamma_i$ that captures domain-specific properties. The intermediate representation $r_i$ is the classifier response $c_i$ subtracted by bridge $\gamma_i$. Similarly, in GVB-D, the bridge layer $D_2$ provides additive discriminating ability to the base domain classifier in discriminator. The network is trained by minimizing classification loss, adversarial loss, discriminator loss and generator loss.}
	\label{softmargin}
\end{figure*}

Visual domain adaptation, first proposed by~\cite{saenko2010adapting}, receives increasing attention and fruitful results~\cite{yang2007cross,wang2012multi}, especially with deep learning techniques~\cite{ganin2014unsupervised,ganin2016domain,haeusser2017associative,hu2018duplex,kang2019contrastive,cui2020nnm}. Techniques developed on deep networks could be mainly categorized into moment alignment and adversarial training. Moment-alignment-based methods~\cite{gretton2012kernel,long2015learning,long2017deep,sun2016deep,yan2017mind} are proposed to measure and minimize the domain discrepancy with maximum mean discrepancy~(MMD)~\cite{long2015learning}, correlation distance~\cite{sun2016deep,zhuo2017deep} or other distance metric~\cite{kang2018deep,long2017deep} calculated on task-specific features.

Motivated by the Generative adversarial networks (GAN) \cite{goodfellow2014generative}, adversarial learning has been successfully explored in various research fields including domain adaptation~\cite{chen2018re,sankaranarayanan2018generate,volpi2018adversarial}.
In Domain Adversarial Neural Network (DANN)~\cite{ganin2016domain}, the authors propose a gradient reversal layer to confuse the  domain classifier. Many methods such as ADDA \cite{tzeng2017adversarial}, CyCADA \cite{hoffman2017cycada}, SBADA~\cite{russo2018source}, and MCD \cite{saito2018maximum} are proposed to build adversarial frameworks to directly minimize the domain discrepancy. Typically, CDAN~\cite{long2018conditional} is a principled framework that conditions the adversarial adaptation models on discriminative information conveyed in the classifier predictions. Symnets~\cite{zhang2019domain} makes a symmetric design of source and target task classifiers, in which both domain classifiers produce classification outputs.

Some methods try to explicitly model the bridge on generator, but they still face many problems. SGF \cite{ni2013subspace} and GFK \cite{gong2012geodesic} build the bridge between domains on Grassmannian manifolds, but could not be easily applied to deep networks. In deep methods, DSN \cite{bousmalis2016domain}, DISE~\cite{chang2019all} and DLOW \cite{gong2019dlow} formulate the bridge between source and target by reconstructing input images on pixel level. However, the time-consuming reconstruction function does not guarantee less residual domain characteristics in the domain-invariant representations. We avoid using image reconstruction function and reduce the range of bridge on generator to get rid of the negative influence of domain-specific characteristics. Besides, multiple discriminators in adversarial networks~\cite{pei2018multi} may break the balance in the adversarial training process with overly critical discriminating ability. In contrast, we pursue a more balanced minmax adversarial game by building the bridge on both generator and discriminator.

\section{Method}

We propose GVB for adversarial domain adaptation as shown in Figure \ref{softmargin}.
In Sec. \ref{ADA}, we introduce the baseline for adversarial domain adaptation. The gradually vanishing bridge on generator (GVB-G) is introduced in Sec. \ref{GVBG}. In Sec. \ref{GVBD}, we describe the gradually vanishing bridge on discriminator (GVB-D) and the whole framework GVB-GD based on GVB-G and GVB-D. 

\subsection{Adversarial Domain Adaptation}
\label{ADA}
We are given source domain $\mathcal{D}_S=\{(x_i^s,y_i^s) \}_{i=1}^{N_s}$ with $N_s$ labeled examples covering $C$ classes where $y_i^s \in \{1,...,C\}$ and target domain $\mathcal{D}_T=\{x_i^t \}_{i=1}^{N_T}$ with $N_t$ unlabeled examples that belong to the same $C$ classes.

It is widely accepted that methods for domain adaptation can be achieved by minimizing a classification loss and an additional transfer loss. The classification loss in $\mathcal{D}_S$ can be calculated as:
\begin{equation}
\mathcal{L}_{cls} = \frac{1}{N_s} \sum_{i=1}^{N_s} L_{ce} (G_{\star}(x_i^s)),y_i^s)
\label{cls} \end{equation}
where $L_{ce}$ is the cross-entropy loss function and $G_{\star}$ is the network structure to obtain class responses. To construct transfer loss, inspired by~\cite{long2015learning,saito2018maximum}, we directly minimize the domain discrepancy on classifier responses. Besides, we adopt popular methods of adversarial training, the same with other deep domain adaptation methods~\cite{ganin2014unsupervised,hoffman2017cycada,long2018conditional}. Thus the transfer loss could be formulated as:
\begin{equation}
\begin{split}{
	\mathcal{L}_{trans}^{adv}=& -\frac{1}{N_s} \sum_{i=1}^{N_s} \log D_{\star}(G_{\star}(x_i^s))) \\
	&- \frac{1}{N_t}\sum_{j=1}^{N_t} \log (1- D_{\star}(G_{\star}(x_j^t)))),
}\end{split}
\label{trans}
\end{equation}
where $D_{\star}$ is the constructed discriminator. Some methods utilize extra loss, such as reconstruction loss, which can be denoted by $\mathcal{L}_{ext}$. The overall objective function is:
\begin{equation}
\begin{split}{
	\quad &\min_{G_{\star}}{\quad	\mathcal{L}_{cls} + \mathcal{L}_{trans}^{adv}} +\mathcal{L}_{ext} \\
	\quad &\max_{D_{\star}}{\quad \mathcal{L}_{trans}^{adv}}. \\
}\end{split}
\label{transadv}
\end{equation}

This formulation is regarded as the general framework for adversarial domain adaptation. In particular, suppose that the classification network consists of a feature extractor $G_1$ and a successive classifier layer $G_2$. For the input data $x_i$, the feature output and the classifier response can be computed as $f_i=G_1(x_i)$ and $c_i=G_2(f_i)$, respectively. With the discriminator $D_1$, the baseline of our method could be expressed as:
\begin{equation} \left\{
\begin{aligned}
&G_{\star}(x_i) = G_2(G_1(x_i)) \\
&D_{\star}(G_{\star}) = D_1(G_{\star}), \quad \mathcal{L}_{ext} = 0,
\label{framework}
\end{aligned}
\right.
\end{equation}

\subsection{Gradually Vanishing Bridge on Generator}
\label{GVBG}

Actually, divergence between source and target domains can hardly be minimized to zero, thus it is quite challenging to seek direct knowledge transfer from source domain $\mathcal{D}_S$ to target domain $\mathcal{D}_T$. Alternatively, to alleviate the transfer difficulty, we divide the transfer process into two separate processes from either source or target domain to intermediate domain $\mathcal{D}_I$. $\mathcal{D}_I$ is supposed to lie in the middle of $\mathcal{D}_S$ and $\mathcal{D}_T$, with much less domain-specific characteristics than $\mathcal{D}_S$ and $\mathcal{D}_T$. To model domain-specific representations, we construct a bridge layer $G_3$ in light blue boxes in Figure~\ref{softmargin}. The output of $G_3$ is the bridge denoted as $\gamma_i$, which captures the domain-specific representation from the input $x_i$ in either $\mathcal{D}_S$ or $\mathcal{D}_T$, {\it i.e.}, $\gamma_i = G_3(G_1(x_i))$. Subtracting bridge $\gamma_i$ from the classifier response $c_i$, results in the constructed domain-invariant representation, denoted as intermediate representation $r_i$, calculated as follows:
\begin{equation}
\begin{split}{
	&r_i = c_i - \gamma_i. \\
}\end{split}
\label{change}
\end{equation}
With bridge $\gamma_i$, the framework is the same as Eqn.~\ref{framework} except the generator. And the overall generator becomes:
\begin{equation}
\begin{split}{
	G_{\star}(x_i)=G_2(G_1(x_i))-G_3(G_1(x_i)). 
}\end{split}
\end{equation}
The adversarial training process enforces that the distributions of $r_i$ across domains are as similar as possible. Therefore, $\gamma_i$ tends to be dominated by domain-specific representations according to~Eqn.~\ref{change}.

In existing study, some bridge-based methods on the generator have been proposed~\cite{bousmalis2016domain,chang2019all,gong2019dlow}, but the ways of constructing the bridge of existing study remain to be discussed. The bridge is built by image reconstruction in~\cite{bousmalis2016domain,chang2019all,gong2019dlow}. 
However, the image reconstruction function inevitably results in too much residual domain-specific characteristics and large overall range in the constructed bridge. In connection to our framework, if the range of $\gamma_i$, measured in vector-norm, is overly large, it has the following interpretations. First, it implies that data $x_i$ is a hard example which contains heavy domain-specific properties. Second, the domain-specific and domain-invariant parts for $x_i$ can hardly be separated. Consequently, the large range of the bridge means rich domain characteristics in the $\gamma_i$, and inevitably impacts $c_i$ and $r_i$. The influence on the representations leads to a higher misclassification probability in $\mathcal{D}_T$, which is validated by experiment results in Figure~\ref{GVB-G}.

\begin{figure}[t]
	\begin{center}
		%#fbox{\rule{0pt}{2in} \rule{0.9\linewidth}{0pt}}
		\includegraphics[width=0.9\linewidth]{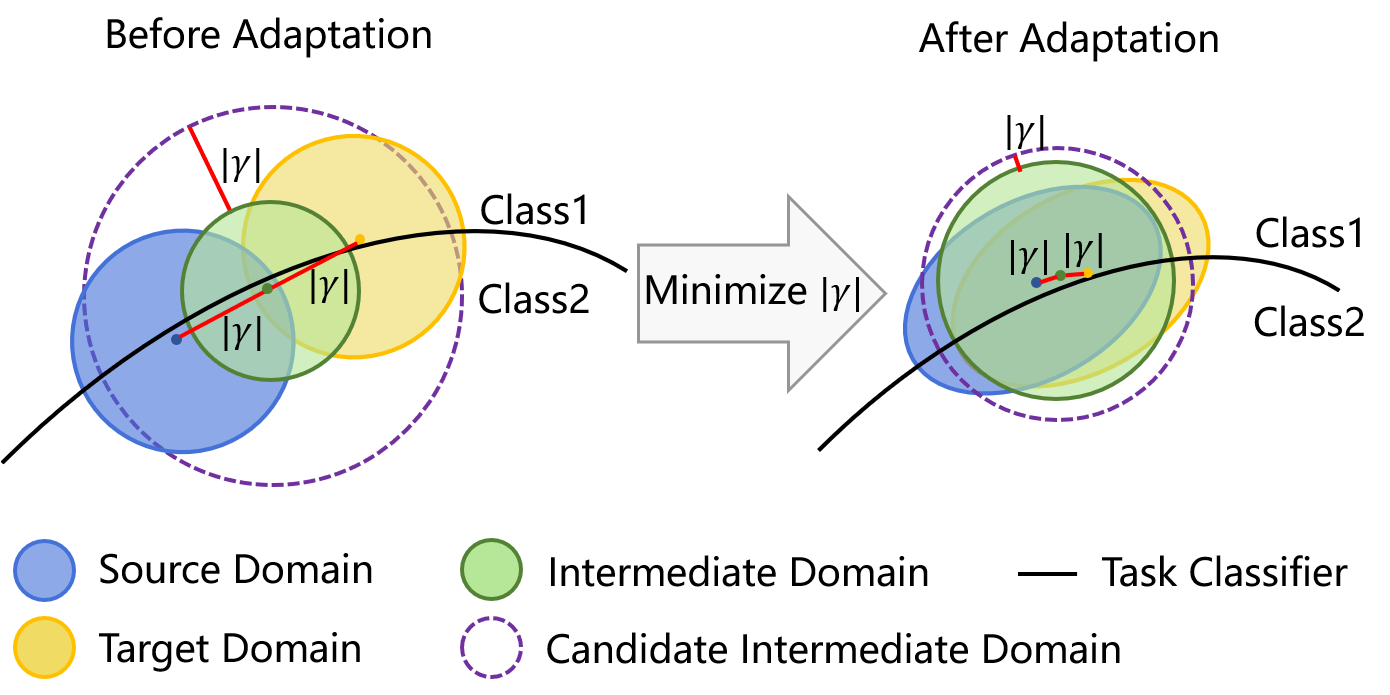}
	\end{center}
	\caption{A toy example of the training process with bridge $\gamma$. With grdually minimized range of $\gamma$, the discrepancy between source and target domains is reduced. After adaptation, more points are covered in intermediate domain. But the points still outside of intermediate domain are regarded as hard examples or noisy data.}
	\label{trainprocess}
\end{figure}

Towards more effective constraint on the bridge, we seek to reduce the influence of domain characteristics in $\gamma_i$. Thus we gradually reduce the influence of $\gamma_i$ by progressively minimizing the overall range of $\gamma_i$ as follows:
\begin{equation}
\begin{split}{
	\mathcal{L}_{G} = \frac{1}{(N_s+N_t)*C}\sum_{i=1}^{N_s+N_t}\sum_{j=1}^{C}|\gamma_{i,j}|,
}\end{split}
\label{minloss}
\end{equation}
where each $\gamma_i$ is encouraged to be near zeroes. With the reduced range of $\gamma_i$, the characteristics in $\gamma_i$ is suppressed, which results in minor negative effects on $c_i$ and $r_i$. The mechanism of minimizing the range of $\gamma_i$ is denoted as GVB. We denote GVB applied to generator as GVB-G. As shown in the experiment in Figure~\ref{examples}, visualization of the data with and without $\gamma_i$ also validates that $\gamma_i$ still contains rich domain characteristics without input reconstruction.

For better comprehension of the gradually vanishing bridge mechanism, we show the example in two-dimension space in Figure \ref{trainprocess} working on generator. 
We denote $\mathcal{D}_S$, $\mathcal{D}_T$ and $\mathcal{D}_I$ with circles in different colors. For simplicity, we assume that the range of bridge $\gamma$ is the single parameter to be optimized, which can measure the distance between the centers of $\mathcal{D}_S$($\mathcal{D}_T$) and $\mathcal{D}_I$. The distance shows the discrepancy between $\mathcal{D}_S$($\mathcal{D}_T$) and $\mathcal{D}_I$, which could approximately measure the quantity of domain-specific characteristics in $\mathcal{D}_S$($\mathcal{D}_T$). $|\gamma|$ appears to be large before domain adaptation, which demonstrates the heavy domain characteristics in $\mathcal{D}_S$($\mathcal{D}_T$). Based on Eqn. \ref{minloss}, $|\gamma|$ is minimized to reduce domain discrepancy. In this way, $\mathcal{D}_S$ and $\mathcal{D}_T$ will be pushed to be closer and overlapped with $\mathcal{D}_I$ as much as possible to obtain domain-invariant characteristics. In this case, there is less domain-specific characteristics in $\mathcal{D}_S$($\mathcal{D}_T$), resulting more reliable domain-invariant representations in $\mathcal{D}_I$.

Furthermore, all the data points, from either $\mathcal{D}_S$ or $\mathcal{D}_T$ that are possibly dragged into $\mathcal{D}_I$ ({\it i.e.}, translated with a distance of less than $|\gamma|$), are within the region bounded by dashed circle, which is referred to as the candidate intermediate domain. In the training process, the points in the candidate intermediate domain will be dragged into $\mathcal{D}_I$. After domain adaptation, $|\gamma|$ becomes small, and the candidate intermediate domain will shrink to near the sphere of $\mathcal{D}_I$. In this case, most of the examples will be included into $\mathcal{D}_I$ ({\it i.e.}, $|\gamma_i|$ is near zero), while the ones in $\mathcal{D}_S$ or $\mathcal{D}_T$ but outside the candidate intermediate domain are recognized as the hard examples that contain too much domain-specific characteristics.

\subsection{Gradually Vanishing Bridge on Discriminator}
\label{GVBD}

In adversarial domain adaptation, less attention has been paid to the discriminator. Besides, unlike the generator, which could be pretrained on large-scale datasets such as ImageNet \cite{deng2009imagenet}, the discriminator is usually randomly initialized. Due to the large computation cost and searching space, the discriminator is prone to run into local minimum. Towards more effective discriminator, we propose to build a bridge layer $D_2$ on the discriminator as shown in dark blue boxes in Figure \ref{softmargin}. $D_2$ models the distance between the basic discriminator $D_1$ function hyperplane and ideal decision boundary, and provides additive discriminating power to $D_1$. Thus the overall discriminator becomes:
\begin{equation}
\begin{split}{
	D_{\star}(G_{\star}(x_i))= D_1(G_{\star}(x_i)) + D_2(G_{\star}(x_i)).
}\end{split}
\label{disadv}
\end{equation}

In the two-player game of adversarial learning, it is a common sense that balanced players could stimulate and improve each other more significantly. For example, multiple discriminators~\cite{pei2018multi} may provide domain discrimination from different aspects, but they should be combined properly for more balanced adversarial learning, {\it e.g.}, combined with a more capable generator. For GVB, the ability of generator is enhanced with the introduction of GVB-G, thus the discriminating ability of the discriminator side should also be adjusted accordingly. Therefore, similar to GVB-G, we progressively minimize the overall range of the bridge on discriminator side as follows:
\begin{equation}
\begin{split}{
	\mathcal{L}_{D} = \frac{1}{(N_s+N_t)}\sum_{i=1}^{N_s+N_t}|\sigma_i|,
}\end{split}
\label{discloss}
\end{equation}
where $\sigma_i$ is the constructed bridge, {\it i.e.}, $\sigma_i=D_2(G_{\star}(x_i))$. We denote GVB applied to the discriminator as GVB-D.

Towards more balanced minmax game in the whole training process, we combine GVB-G and GVB-D, and formulate the gradually vanishing bridge on generator and discriminator (GVB-GD). The whole framework could be described as follows:
\begin{equation} \left\{
\begin{aligned}
&G_{\star}(x_i) = G_2(G_1(x_i)) - G_3(G_1(x_i)) \\
&D_{\star}(G_{\star}) = D_1(G_{\star}) - D_2(G_{\star}) \\
&\mathcal{L}_{ext} = \lambda \mathcal{L}_{G} + \mu \mathcal{L}_{D},
\end{aligned}
\right.
\end{equation}
where the extra loss combines both Eqn. \ref{minloss} and Eqn. \ref{discloss} with hyper-parameters $\lambda$ and $\mu$. Since our proposed GVB mechanism focuses on optimization of minmax game, it is general applicable to adversarial domain adaptation framework. Thus we also apply GVB to existing methods including CDAN~\cite{long2018conditional} and Symnets~\cite{zhang2019domain}. CDAN multiplies the feature representations and class responses on the generator, to play against the discriminator. Both GVB-G and GVB-D could be applied to CDAN. Symnets builds symmetric generators for domains, but the discriminator is not modeled explicitly. Therefore, only GVB-G could be applied to SymNets.

\section{Experiment}
\begin{table*}[htbp]
	\begin{center}
		
		%\vspace{-12pt}
		
		\caption{Accuracies (\%) on Office-31 for ResNet50-based unsupervised domain adaptation methods.}
		\vspace{5pt}
		
		\label{tableoffice}
		\addtolength{\tabcolsep}{2pt}
		\scalebox{0.92}{
			\resizebox{\textwidth}{!}{%
				\begin{tabular}{cccccccc}
					\hline
					Method & A $\rightarrow$ D & A $\rightarrow$ W & D $\rightarrow$ W & W $\rightarrow$ D & D $\rightarrow$ A & W $\rightarrow$ A & Avg \\
					\hline\hline
					
					ResNet-50 \cite{he2016deep} & 68.9$\pm$0.2 & 68.4$\pm$0.2 & 96.7$\pm$0.1 & 99.3$\pm$0.1 & 62.5$\pm$0.3 & 60.7$\pm$0.3 & 76.1 \\
					GFK \cite{gong2012geodesic} & 74.5$\pm$0.0 & 72.8$\pm$0.0 & 95.0$\pm$0.0 & 98.2$\pm$0.0 & 63.4$\pm$0.0 & 61.0$\pm$0.0 & 77.5 \\
					DAN \cite{long2015learning} & 78.6$\pm$0.2 & 80.5$\pm$0.4 & 97.1$\pm$0.2 & 99.6$\pm$0.1 & 63.6$\pm$0.3 & 62.8$\pm$0.2 & 80.4 \\
					DANN \cite{ganin2016domain} & 79.7$\pm$0.4 & 82.0$\pm$0.4 & 96.9$\pm$0.2 & 99.1$\pm$0.1 & 68.2$\pm$0.4 & 67.4$\pm$0.5 & 82.2 \\
					ADDA \cite{tzeng2017adversarial} & 77.8$\pm$0.3 & 86.2$\pm$0.5 & 96.2$\pm$0.3 & 98.4$\pm$0.3 & 69.5$\pm$0.4 & 68.9$\pm$0.5 & 82.9 \\
					Simnet \cite{pinheiro2018unsupervised} & 85.3$\pm$0.3 & 88.6$\pm$0.5 & 98.2$\pm$0.2 & 99.7$\pm$0.2 & {73.4}$\pm$0.8 & 71.8$\pm$0.6 & 86.2 \\				
					GTA \cite{sankaranarayanan2018generate} & 87.7$\pm$0.5 & 89.5$\pm$0.5 & 97.9$\pm$0.3 & 99.8$\pm$0.4 & 72.8$\pm$0.3 & 71.4$\pm$0.4 & 86.5 \\
					MCD \cite{saito2018maximum} & {92.2}$\pm$0.2 & {88.6}$\pm$0.2 & {98.5}$\pm$0.1 & \textbf{100.0}$\pm$.0 & {69.5}$\pm$0.1 & {69.7}$\pm$0.3 & {86.5} \\
					MDD \cite{zhang2019bridging} & {93.5}$\pm$0.2 & {94.5}$\pm$0.3 & {98.4}$\pm$0.1 & \textbf{100.0}$\pm$.0 & {74.6}$\pm$0.3 & {72.2}$\pm$0.1 & {88.9} \\	
					\hline	
					
					Baseline & 89.9$\pm$0.3 & 92.5$\pm$0.5 & 98.5$\pm$0.3 & 99.9$\pm$0.1 & 70.0$\pm$0.4 & 71.1$\pm$0.3 & 87.0 \\
					GVB-G & {93.9}$\pm$0.4 & {94.2}$\pm$0.4 & {98.6}$\pm$0.2 & \textbf{100.0}$\pm$.0 & {71.8}$\pm$0.3 & {73.5}$\pm$0.4 & {88.7} \\
					GVB-D & {92.8}$\pm$0.5 & {93.9}$\pm$0.4 & {98.4}$\pm$0.2 & \textbf{100.0}$\pm$.0 & {72.0}$\pm$0.3 & {72.6}$\pm$0.3 & {88.3} \\
					GVB-GD & {95.0}$\pm$0.4 & \textbf{94.8}$\pm$0.5 & {98.7}$\pm$0.3 & \textbf{100.0}$\pm$.0 & {73.4}$\pm$0.3 & \textbf{73.7}$\pm$0.4 & {89.3} \\
					\hline
					CDAN \cite{long2018conditional} & {92.9}$\pm$0.2 & {93.1}$\pm$0.1 & {98.6}$\pm$0.1 & \textbf{100.0}$\pm$.0 & {71.0}$\pm$0.3 & {69.3}$\pm$0.3 & {87.5} \\
					CDAN-G & {92.1}$\pm$0.3 & {92.9}$\pm$0.2 & {98.2}$\pm$0.1 & \textbf{100.0}$\pm$.0 & {73.5}$\pm$0.3 & {72.8}$\pm$0.2 & {88.2} \\
					CDAN-D & {93.5}$\pm$0.4 & {92.9}$\pm$0.2 & {98.6}$\pm$0.1 & \textbf{100.0}$\pm$.0 & {73.0}$\pm$0.3 & {73.1}$\pm$0.3 & {88.5} \\
					CDAN-GD & {93.7}$\pm$0.2 & {94.0}$\pm$0.2 & {98.6}$\pm$0.1 & \textbf{100.0}$\pm$.0 & {73.4}$\pm$0.3 & {73.0}$\pm$0.2 & {88.8} \\
					\hline
					Symnets \cite{zhang2019domain} & {93.9}$\pm$0.5 & {90.8}$\pm$0.1 & \textbf{98.8}$\pm$0.3 & \textbf{100.0}$\pm$.0 & {74.6}$\pm$0.6 & {72.5}$\pm$0.5 & {88.4} \\
					Symnets-G  & \textbf{96.1}$\pm$0.3 & {93.8}$\pm$0.4 & \textbf{98.8}$\pm$0.2 & \textbf{100.0}$\pm$.0 & \textbf{74.9}$\pm$0.4 & {72.8}$\pm$0.3 & \textbf{89.4} \\
					\hline
			\end{tabular}}
			
		}
	\end{center}
	\vspace{0pt}
\end{table*}
%TCA \cite{pan2011domain} & 74.1$\pm$0.0 & 72.7$\pm$0.0 & 96.7$\pm$0.0 & 99.6$\pm$0.0 & 61.7$\pm$0.0 & 60.9$\pm$0.0 & 77.6 \\
\begin{table*}[htbp]
	\begin{center}
		\vspace{-2pt}
		
		\caption{Accuracies (\%) on Office-Home for ResNet50-based unsupervised domain adaptation methods.}
		\vspace{-5pt}
		\label{tableofficehome}
		\addtolength{\tabcolsep}{-5.5pt}
		\resizebox{\textwidth}{!}{%
			\begin{tabular}{cccccccccccccc}
				\hline
				Method& Ar$\rightarrow$Cl & Ar$\rightarrow$Pr & Ar$\rightarrow$Rw & Cl$\rightarrow$Ar & Cl$\rightarrow$Pr & Cl$\rightarrow$Rw & Pr$\rightarrow$Ar & Pr$\rightarrow$Cl & Pr$\rightarrow$Rw & Rw$\rightarrow$Ar & Rw$\rightarrow$Cl & Rw$\rightarrow$Pr & Avg \\
				\hline\hline
				
				ResNet-50 \cite{he2016deep} & 34.9 & 50.0 & 58.0 & 37.4 & 41.9 & 46.2 & 38.5 & 31.2 & 60.4 & 53.9 & 41.2 & 59.9 & 46.1 \\
				DAN \cite{long2015learning} & 43.6 & 57.0 & 67.9 & 45.8 & 56.5 & 60.4 & 44.0 & 43.6 & 67.7 & 63.1 & 51.5 & 74.3 & 56.3 \\
				DANN \cite{ganin2016domain} & 45.6 & 59.3 & 70.1 & 47.0 & 58.5 & 60.9 & 46.1 & 43.7 & 68.5 & 63.2 & 51.8 & 76.8 & 57.6 \\
				MCD \cite{saito2018maximum} &48.9	&68.3	&74.6	&61.3	&67.6	&68.8	&57	&47.1	&75.1	&69.1	&52.2	&79.6	&64.1 \\
				BNM \cite{cui2020nnm} &{52.3}	&{73.9}	&\textbf{80.0}	&63.3	&{72.9}	&{74.9}	&61.7	&{49.5}	&{79.7}	&70.5	&{53.6}	&82.2	&67.9\\
				MDD \cite{zhang2019bridging} & 54.9 & 73.7 & 77.8 & 60.0 & 71.4 & 71.8 & 61.2 & 53.6 & 78.1 & 72.5 & \textbf{60.2} & 82.3 & 68.1 \\
				\hline
				Symnets \cite{zhang2019domain} & 47.7 & 72.9 & 78.5 & 64.2 & 71.3 & 74.2 & 63.6  & 47.6 & 79.4 & 73.8 & 50.8 & {82.6} & 67.2 \\
				Symnets-G  & 48.0 & 74.3 & 78.5 & \textbf{65.1} & 72.2 & 74.4 & 65.1  & 49.4 & 79.7 & 73.8 & 51.7 & {82.5} & 67.8 \\
				\hline				
				CDAN \cite{long2018conditional} & 50.7 & 70.6 & 76.0 & 57.6 & 70.0 & 70.0 & 57.4 & 50.9 & 77.3 & 70.9 & 56.7 & {81.6} & 65.8 \\
				CDAN-G  & 53.6 & 72.1 & 77.9 & 61.9 & 71.8 & 71.7 & 63.8 & 52.9 & 79.9 & 71.5 & 58.0 & {83.3} & 68.2 \\
				CDAN-D  & 54.4 & 72.9 & 77.7 & 62.6 & 71.9 & 71.8 & 62.4 & 52.1 & 79.5 & 72.1 & 58.1 & {83.3} & 68.2 \\
				CDAN-GD & 55.3 & 74.1 & 78.2 & 62.4 & 72.6 & 71.8 & {63.8} & 54.1 & 80.1 & 73.1 & 58.7 & 83.6 & 69.0 \\				
				\hline
				Baseline & 54.7 & 72.8 & 78.5 & 62.3 & 71.1 & 73.1 & 61.0 & 53.0 & 80.0 & 72.8 &56.5 & 83.4 &68.3 \\
				GVB-G & {56.5} & {74.0} & {79.2} & {64.2} & {73.0} & {74.1} & \textbf{65.2} & \textbf{55.9} & \textbf{81.2} & {74.2} & {58.2} & {84.0} & {70.0} \\
				GVB-D & {55.0} & {73.8} & {79.0} & {64.3} & {72.9} & \textbf{75.0} & {63.4} & {54.2} & {80.9} & {73.1} & {58.0} & {83.6} & {69.4} \\
				GVB-GD & \textbf{57.0} & \textbf{74.7} & \textbf{79.8} & {64.6} & \textbf{74.1} & {74.6} & \textbf{65.2} & {55.1} & {81.0} & \textbf{74.6} & {59.7} & \textbf{84.3} & \textbf{70.4} \\
				
				\hline
		\end{tabular}}
		
	\end{center}
	\vspace{-5pt}
\end{table*}

We evaluate the GVB mechanism on three challenging standard benchmarks and compare GVB with state-of-the-art domain adaptation methods.
\subsection{Datasets and Settings}
Office-31 \cite{saenko2010adapting} is a standard benchmark for visual domain adaptation which contains 4,652 images in 31 categories. It consists of three domains: Amazon (A), Webcam (W) and DSLR (D). We evaluate the methods across the three domains, {\it i.e.}, six transfer tasks in total.

Office-Home \cite{venkateswara2017deep} is a challenging dataset for visual domain adaptation with 15,500 images in 65 categories. It has four significantly different domains: Artistic images (Ar), Clip Art (Cl), Product images (Pr), and Real-World images (Rw). Among four domains, there are totally 12 challenging domain adaptation tasks. 

VisDA-2017 \cite{visda2017} is a simulation-to-real dataset for domain adaptation with over 280,000 images across 12 categories in the training, validation and testing domains. The training images are generated from the same object under different circumstances, while the validation images are collected from MSCOCO. \cite{lin2014microsoft}. 

\begin{figure*}
	\begin{center}
		%\fbox{\rule{0pt}{2in} \rule{.9\linewidth}{0pt}}
		\includegraphics[width=.97\textwidth]{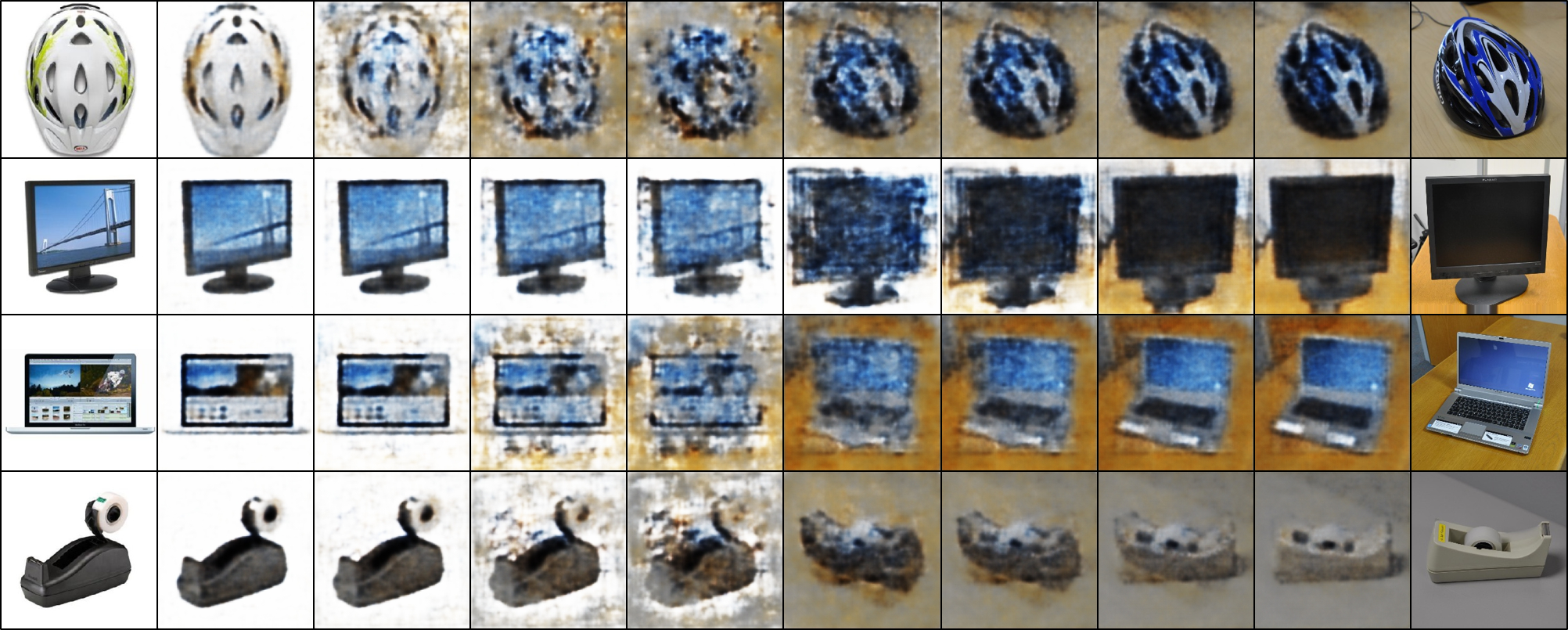}
	\end{center}
	\vspace{-1ex}
	\caption{Visualization of data with and without the bridge $\gamma_i$ on generator on A $\rightarrow$ D for GVB-GD. The images shown in the first and last columns are respectively sampled from source domain (Amazon) and target domain (DSLR). The 2nd$\sim$5th columns show the generated source images corresponding to different representations by subtracting $\gamma_i$ with $0\times, 5\times, 10\times$ and $15\times$ amplifications from classifier responses $c_i$. Similarly, the 6th$\sim$9th columns show the generated target images by subtracting $\gamma_i$ with $15\times, 10\times, 5\times$ and $0\times$ amplifications from $c_i$.}
	\label{examples}
	\vspace{-10pt}
\end{figure*}

For GVB-GD, we adopt ResNet-50 \cite{he2016deep} pre-trained on ImageNet \cite{deng2009imagenet} as our backbone network. We build the bridge layer on both the generator and discriminator with several fully-connected layers in experiments. All the experiments are implemented with PyTorch. For each method, we run four random experiments and report the average classification accuracy and we report the standard deviation in Office-31. We compare our method with state-of-the-art domain adaptation methods. For fair comparison, results on Office-31, Office-Home and VisDA-2017 are directly reported from their original papers if available. Similar to our methods, for CDAN and Symnet, when applying GVB on the generator, `-G' is added to the method name, and `-D' is added when GVB is applied on the discriminator. We compare our methods with state-of-art domain adaptation methods: DAN~\cite{long2015learning}, DANN~\cite{ganin2016domain}, ADDA~\cite{tzeng2017adversarial}, GTA~\cite{sankaranarayanan2018generate}, CDAN~\cite{long2018conditional}, Simnet~\cite{pinheiro2018unsupervised}, MCD~\cite{saito2018maximum} and MDD \cite{zhang2019bridging}.

In the adversarial training, we apply gradient reversal layer~(GRL)~\cite{ganin2014unsupervised} to the network. In GRL, instead of fixing the adaptation factor, progressive training strategy is adopted to suppress noisy signal from the discriminator at the early stages of the training procedure. To optimize the transfer process, we utilize entropy conditioning proposed in \cite{long2018conditional} that pays more attention to easy-to-classify data. We choose the baseline model as described in Sec.~\ref{ADA}. For the training of the network, we employ mini-batch stochastic gradient descent (SGD) with momentum of 0.9 to train our model. We keep fixing $\lambda = 1$ and $\mu = 1$ to balance the loss, since our method of GVB is stable under different values of parameters $\lambda$ and $\mu$.

\begin{table}%[htbp]
	
	\caption{Accuracies (\%) of Synthetic $\rightarrow$ Real on VisDA-2017 for unsupervised domain adaptation methods using ResNet-50.}
	\begin{center}
		\vspace{-15pt}
		\scalebox{1.05}{
			\setlength{\tabcolsep}{1.2mm}{
				\begin{tabular}{cc|cc|cc}%|cccc|cccc
					\hline
					Method &Acc &Method &Acc &Method &Acc\\
					\hline
					DAN \cite{long2015learning} & 61.6 &CDAN \cite{long2018conditional} &70.0 &Baseline&71.3\\
					DANN \cite{ganin2016domain} &57.4 &CDAN-G &73.4&GVB-G&73.1\\
					GTA \cite{sankaranarayanan2018generate} &69.5 &CDAN-D &73.8&GVB-D&72.8\\
					MDD \cite{zhang2019bridging} &74.6 &CDAN-GD &74.9&GVB-GD&\textbf{75.3}\\
					\hline
				\end{tabular}
		}}
	\end{center}
	\label{visda}
	\vspace{-10pt}
\end{table}
\subsection{Results}
\label{vector}

\begin{figure*}
	\begin{center}
		\begin{minipage}{.42\textwidth}
			\begin{minipage}{.46\textwidth}
				
				\subfigure[GVB-G]{	
					\includegraphics[width=0.99\textwidth]{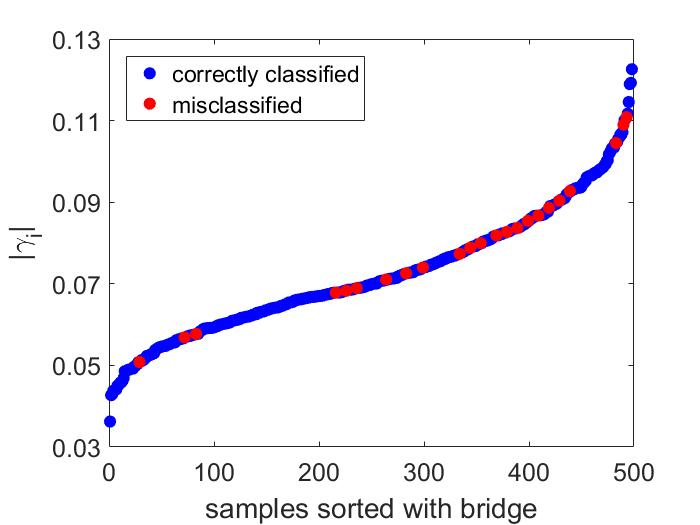}
					\label{GVB-G}}
			\end{minipage}
			\begin{minipage}{.46\textwidth}
				\subfigure[GVB-D]{
					\includegraphics[width=0.99\textwidth]{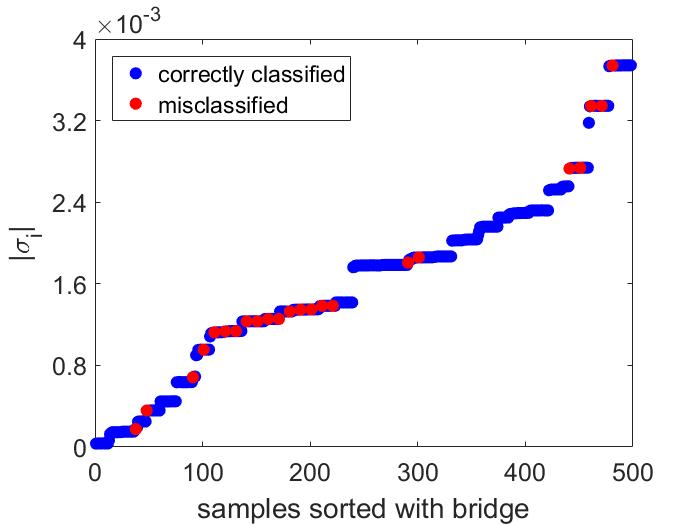}
					\label{GVB-D}}	
			\end{minipage}
			\caption{Relationship between the range of bridge on target domain and classification result on GVB-GD.}
			\label{bridgevector}
		\end{minipage}
		~~~~~~~~
		\begin{minipage}{.5\textwidth}
			\begin{minipage}{.32\textwidth}
				\subfigure[ResNet50]{	
					\includegraphics[width=0.9\textwidth]{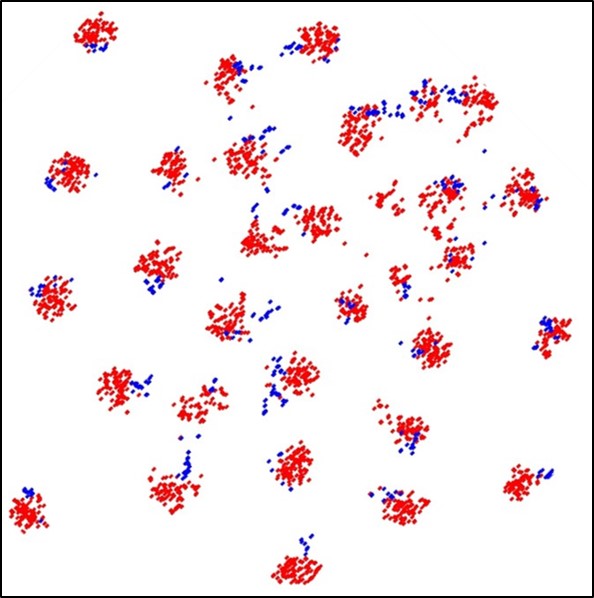}
					\label{tsne1}}
			\end{minipage}
			\begin{minipage}{.32\textwidth}
				\subfigure[Baseline]{
					\includegraphics[width=0.9\textwidth]{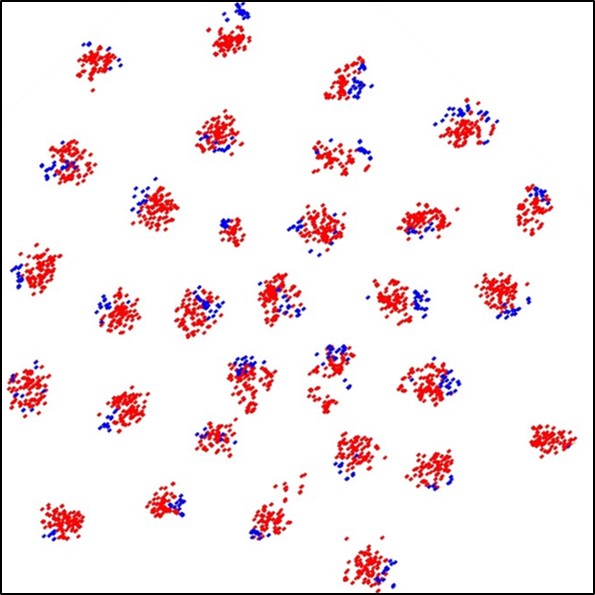}
					\label{tsne2}}
			\end{minipage}
			\begin{minipage}{.32\textwidth}
				\subfigure[GVB-GD]{
					\includegraphics[width=0.9\textwidth]{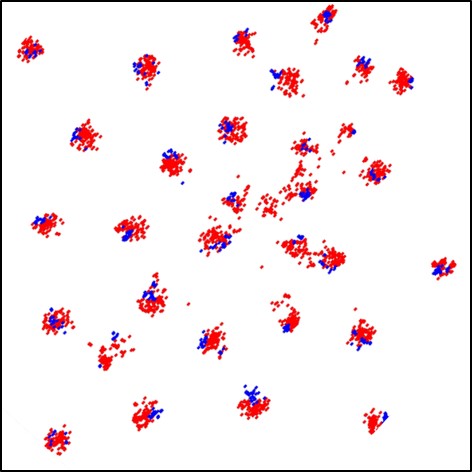}
					\label{tsne4}}
			\end{minipage}

			\caption{$t$-SNE of classifier responses by ResNet50, Baseline and GVB-GD (red: Amazon, blue: Webcam).}
			\label{tsne}
		\end{minipage}
	\end{center}
	\vspace{-2ex}
\end{figure*}

The results on Office-31, Office-Home and VisDA-2017 datasets are separately shown in Table~\ref{tableoffice}, Table~\ref{tableofficehome} and Table~\ref{visda}. Both GVB-G and GVB-D outperform the baseline and most of other competitors significantly, which means the bridge provides prominent improvement for domain adaptation. GVB-GD further outperforms GVB-G and GVB-D for most tasks, which means the bridge on the generator and discriminator can boost each other to learn better domain-invariant representation. Besides, we apply GVB-G to CDAN and Symnets, and GVB-D to CDAN. The well-performed results of CDAN-G, CDAN-D, CDAN-GD and Symnets-G mean the GVB mechanism is generally applicable for mainstream adversarial domain adaptation methods. 

For different datasets on Office-31, Office-Home and VisDA-2017, methods with GVB could achieve well-performed results. Among the datasets, there exists large domain discrepancy in Office-Home and VisDA-2017, while less domain discrepancy in Office. Thus our methods with GVB are suitable for both the easy and difficult tasks. Besides, VisDA-2017 is a large-scale dataset, while Office-Home and Office are relative small datasets. This means our GVB methods are applicable to both small and large datasets. Meanwhile, GVB could cooperate well with methods such as CDAN and Symnets for different tasks. This shows that GVB is generally applicable for a wide range of domain adaptation tasks.

\subsection{Qualitative Validation}

\textbf{Representation visualization.} Since we avoid using image reconstruction in the bridge $\gamma$ on generator, there remain concerns on whether there are rich domain-specific characteristics in $\gamma$. We visualize the representations with and without $\gamma$ in Figure \ref{examples} on Office-31. 500 random images from Amazon and 498 images from DSLR are selected to train the network. Similar to FCN \cite{long2015fully}, we use the encoder-decoder network to achieve the input images reconstruction. The encoder is the same as GVB-GD and the decoder is a deconvolutional neural network that maps classifier responses into images. We utilize the weights of encoder with pre-trained GVB-GD on A $\rightarrow$ D and fix the encoder parameters in the training process. We train the encoder-decoder network to reconstruct images from both domains by minimizing $L_2$ loss between input image and reconstructed image. As the range of $\gamma$ is small compared with classification responses $c$, it is difficult to directly visualize $\gamma$. Thus we amplify the $\gamma$ and construct several representations $r = c - \sigma \gamma$ with $\sigma \in \{0, 5, 10, 15\}$ as the inputs of decoder for better visualization.

From the left to right, we can see the gradual progress from source domain to target domain in Figure \ref{examples}. Take the first row for example, the body of the bike helmet is white in source domain and blue in target domain. With increasing amplification factor $\sigma$, the generated helmets from the 2nd to the 5th column become darker and more blue, while those from the 9th to the 6th column becomes whiter. Besides, the helmet is vertical in source domain and slant in target domain, while the object region of the generated images with amplified $\gamma$ is in a circle shape which covers the overlapping region of the original images. Furthermore, the background is white in source domain and brown in target domain, and it appears as half white and half brown in the images with amplified $\gamma$. All these results validate that $\gamma$ captures rich domain-specific representations.

\begin{table}%[htbp]
	%TODO
	\caption{Accuracies (\%) on Office-Home with ResNet50. Methods with notion of BG (D) are the bridge without minimization.}
	\begin{center}
		\vspace{-15pt}
		\scalebox{1.0}{
			\setlength{\tabcolsep}{1.2mm}{
				\begin{tabular}{cccccc}%|cccc|cccc
					\hline
					Method& Ar$\rightarrow$Cl & Ar$\rightarrow$Pr & Ar$\rightarrow$Rw & Cl$\rightarrow$Ar &Avg\\
					\hline
					Baseline & 54.7 & 72.8 & 78.5 & 62.3& 67.1\\
					\hline
					BG &54.1	&72.3 	&78.7	&62.9	&67.0\\
					GVB-G & {56.5} & {74.0} & {79.2} & {64.2} &68.5\\
					\hline
					BD&55.3	&73.1	&79.0	&63.9	&67.8\\	
					GVB-D & {55.0} & {73.8} & {79.0} & {64.3} &68.0\\
					\hline
					GVB-G+BD &56.2	&74.4	&79.6&	64.2	&68.6 \\				
					GVB-GD & \textbf{57.0} & \textbf{74.7} & \textbf{80.0} &\textbf{64.8} &\textbf{69.1}\\
					\hline
				\end{tabular}
		}}
	\end{center}
	\label{Ablation}
	\vspace{-10pt}
\end{table}
\textbf{Statistical analysis on the bridge.}
To analyze the relationship between the bridge on target domain and classification results, we calculate the range of the bridge as $|\gamma_i|$ on GVB-GD. In Figure~\ref{bridgevector}, we show the results of A $\rightarrow$ D on DSLR and sort samples according to the range of bridge. In Figure~\ref{GVB-G}, with increasing range of the bridge $\gamma_i$ on generator, the red dots become more intensive and the classification error goes higher. This validates that higher misclassification probability appears with larger $|\gamma_i|$, thus minimizing the range of $\gamma_i$ tends to result in lower misclassification probability. 

On the discriminator side, as observed in Figure~\ref{GVB-D}, $\sigma_i$ tends to be large when the input representations could be easily transformed into intermediate domain, {\it i.e.}, they tend to be more domain-invariant. These ``easy" examples need more discriminative ability from discriminator to distinguish which domain they come from, and they need less efforts on generator side to produce domain-invariant representations, resulting smaller $|\gamma_i|$. More importantly, these ``easy" samples tend to be easily classified with higher probability since they show more domain-invariance.
By contrast, samples with larger $|\gamma_i|$ deliver more domain-specific properties, which results in smaller $|\sigma_i|$ and higher misclassification probability.

\textbf{Ablation study.}
To validate the effect of GVB compared with the single bridge, we conduct ablation experiments as shown in Table~\ref{Ablation}.  Methods with notion of B-G and G-D are the bridge without minimization, {\it i.e.}, $\lambda=0$ and $\mu=0$. (GVB-G)+(B-D) means to add a bridge on discriminator without minimization to replace the GVB-D, {\it i.e.}, $\mu=0$. GVB-G outperforms B-G by a large margin, this means the gradually vanishing mechanism is indispensible on the generator. Results on GVB-D and B-D are similar, but GVB-GD outperforms (GVB)-G+(B-D). This means GVB-D could cooperate with GVB-G better than B-D for a more balanced adversarial training.

\textbf{Feature visualization.}
We visualize the features of ResNet50 and our methods on A $\rightarrow$ D by $t$-SNE \cite{maaten2008visualizing} in Figure \ref{tsne}. It is shown in Figure \ref{tsne1} that ResNet50 works well in source domain (Amazon) but poorly in target domain (Dslr). The Baseline achieves well-performed global alignment by transferring on classifier responses as shown in Figure \ref{tsne2}, but the direct transfer leads to a large average distance between target data and its nearest source data. For GVB-GD, target samples are located closer to the source examples. The clusters in Figure \ref{tsne4} are more compact that the other two cases, which further validates the learning ability of our GVB mechanism.

\section{Conclusion}
In this paper, we propose the gradually vanishing bridge mechanism that can work on the generator and discriminator in adversarial domain adaptation. On the generator, GVB could result in more domain-invariant representation and better reduction of the negative influence of the rich domain-specific characteristics. To achieve a more balanced adversarial training process, GVB is also built on the discriminator side to provide additive discrimination power. We also apply GVB to existing methods and achieve remarkable improvement over their original counterparts.

\vspace{3pt}
\textbf{Acknowledgement}. This work was supported in part by the National Key R\&D Program of China under Grant 2018AAA0102003, in part by National Natural Science Foundation of China: 61672497, 61620106009, 61836002, 61931008 and U1636214, and in part by Key Research Program of Frontier Sciences, CAS: QYZDJ-SSW-SYS013. Authors would like to thank Kingsoft Cloud for support of free GPU cloud computing resource and helpful disscussion.

{\small
\bibliographystyle{ieee_fullname}
\bibliography{egbib}
}

\end{document}